\documentclass[lettersize,journal]{IEEEtran}
\usepackage{lineno,hyperref}
\usepackage{amsfonts, mathtools, amsmath}
\usepackage{xcolor,color,soul}
\usepackage{multirow} 
\usepackage{multicol} 
\usepackage{makecell}
\usepackage{diagbox}
\usepackage{arydshln}
\usepackage{url}

\usepackage[OT1]{fontenc}

\usepackage{array}
\usepackage{booktabs}

\soulregister{\cite}7
\soulregister{\ref}7

\begin{document}

\title{AGPCNet: Attention-Guided Pyramid Context Networks for Infrared Small Target Detection}

\author{Tianfang~Zhang\href{https://orcid.org/0000-0003-4183-7053}{\includegraphics[scale=0.08]{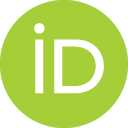}},~Siying~Cao\href{https://orcid.org/0000-0003-1488-6773}{\includegraphics[scale=0.08]{./pics/ORCIDiD.png}},~Tian~Pu\href{https://orcid.org/0000-0002-4419-4041}{\includegraphics[scale=0.08]{./pics/ORCIDiD.png}},~Zhenming~Peng\href{https://orcid.org/0000-0002-4148-3331}{\includegraphics[scale=0.08]{./pics/ORCIDiD.png}},~\IEEEmembership{Member,~IEEE}
\thanks{This work is supported by National Natural Science Foundation of China (61775030, 61571096) and Sichuan Science and Technology Program (2019YJ0167).\textit{(Corresponding author: Zhenming Peng and Tian Pu)}}
\thanks{Tianfang Zhang, Siying Cao, Tian Pu and Zhenming Peng are with the Laboratory of Imaging Detection and Intelligent Perception, School of Information and Communication Engineering, University of Electronic Science and Technology of China, Chengdu 610054, China (e-mail: sparkcarleton@gmail, caosiying3008@gmail.com, putian@uestc.edu.cn, zmpeng@uestc.edu.cn)} 
}

\markboth{Journal of \LaTeX\ Class Files,~Vol.~14, No.~8, August~2015}%
{Shell \MakeLowercase{\textit{et al.}}: Bare Demo of IEEEtran.cls for IEEE Journals}
%



\maketitle

\begin{abstract}
Infrared small target detection is an important problem in many fields such as earth observation, military reconnaissance, disaster relief, and has received widespread attention recently. This paper presents the Attention-Guided Pyramid Context Network (AGPCNet) algorithm. Its main components are an Attention-Guided Context Block (AGCB), a Context Pyramid Module (CPM), and an Asymmetric Fusion Module (AFM). AGCB divides the feature map into patches to compute local associations and uses Global Context Attention (GCA) to compute global associations between semantics, CPM integrates features from multi-scale AGCBs, and AFM integrates low-level and deep-level semantics from a feature-fusion perspective to enhance the utilization of features. The experimental results illustrate that AGPCNet has achieved new state-of-the-art performance on two available infrared small target datasets. The source codes are available at \url{https://github.com/Tianfang-Zhang/AGPCNet}.
\end{abstract}


%
\IEEEpeerreviewmaketitle

\section{Introduction}

\IEEEPARstart{A}{t} present, infrared search and track systems (IRST) are widely used in many fields such as military early warning, marine rescue, and forest fire prevention\cite{huang2020structure}. Compared with other imaging methods, IRST has the advantages of long imaging distance, clear images, and anti-interference. However, also due to these characteristics, the detected targets are very small and lack shape, color, and texture information. At the same time, there are complex clutter and random noise interference in the background. These characteristics make infrared small target detection a challenging task, while it has received widespread attention\cite{wang2017infrared}.

In the past decades, infrared small target detection methods have been model-driven due to the lack of public datasets. The model-driven approaches focus on the physical characteristics of targets, analyze their imaging characteristics on the detector, and designs detection algorithms for different hypotheses. Background suppression-based approaches assume that the infrared image is continuously changing and the presence of small targets breaks the original continuity\cite{tom1993morphology, deshpande1999max}. Human visual system-based approaches assume that humans are able to distinguish small targets based on the difference between the local extent of targets and background\cite{chen2013local, wei2016multiscale, bai2018derivative, shi2017high}. From another perspective, optimization-based approaches consider matrix perspective. They regard an infrared image as a linear superposition of target image and background image\cite{gao2013infrared, lin2009fast, wang2017infraredtvpcp}. By analyzing the properties of two matrices and applying different constraints, the target detection problem is transformed into a matrix decomposition optimization problem\cite{zhang2018infrarednram, zhang2019infrarednolc, zhang2021infraredsrws}. Constrained by the fact that the hypotheses of model-driven approaches do not satisfy all the complexities, these approaches are always inadequate for the actual detection task. Despite the proliferation of theories and constraint items, they are still plagued by complex parameters, low detection accuracy, and low robustness\cite{dai2017reweighted, zhang2019infraredpstnn, kong2021infrared}.

Recently, with the availability of public datasets\cite{wang2019miss, dai2021asymmetric}, data-driven approaches are attracting more and more attention. Compared to previous approaches, in terms of detection proposes, model-driven tend to detect the presence of targets at a certain location rather than segmenting or detecting target in its entirety, whereas data-driven hopes to obtain accurate segmentation of small targets through pixel-level annotation of the datasets. In terms of detection process, model-driven relies on reasonable assumptions and does not require a training process, whereas data-driven requires large amounts of data to train detectors. In terms of detection results, data-driven approaches have more advantages in terms of both positive detection and false alarms\cite{wang2019miss, dai2021asymmetric}.

In the following we will take a more detailed look at data-driven approaches. MDvsFA cGan\cite{wang2019miss} divides the generator into two subtasks, miss detection and false alarm, and proposed a generative adversarial network. On the other hand, ACM\cite{dai2021asymmetric} focuses on feature fusion. It obtains feature maps through the encoder-decoder structure and then fuses the low-level and deep-level semantics using an asymmetric structure based on the characteristics of the information contained in them, obtaining a more efficient feature representation. Although they all approach the problem of infrared small target detection from different perspectives, they still have many shortcomings. Firstly, the stacked convolution operation limits the perceptual field of networks, whereas we usually need global information to discriminate targets position. On the other hand, some methods are limited to single-scale measurements in the acquisition of global information, which also limits and the detection accuracy of networks. In addition, the feature fusion process constrains the low-level and deep-level semantics separately can create the problem of feature mismatch and reduce the feature representation capability of networks.

In order to solve these problems, we propose a learning-based and data-driven scheme called Attention-Guided Pyramid Context Network (AGPCNet). For global information acquisition, we propose Attention-Guided Context Block (AGCB), which divides the feature map into patches to compute the local correlation of features, and then computes the global correlation between patches by Global Context Attention (GCA) to obtain the global information between pixels. For multi-scale feature acquisition, we propose Context Pyramid Module (CPM), which fuses multiple scales of AGCBs with the original feature map to obtain a more accurate feature representation. For feature fusion, we propose Asymmetric Fusion Module (AFM), which performs asymmetric feature filtering after fusion to solve the mismatch problem.

To validate the effectiveness of the proposed AGPCNet, we conducted extensive ablation studies for backbone, CPM, AFM and so on. In order to make SIRST\cite{dai2021asymmetric} dataset more suitable for training neural networks, it has been augmented and published as SIRST Aug. The comparison experiments with state-of-the-art methods were conducted on both MDFA\cite{wang2019miss} and SIRST Aug datasets. In addition, we unified the image inputs of both datasets into $256 \times 256$ closer to the actual infrared scene and merged the two datasets. Comparative experiments were also carried out on the merged dataset. The experimental results demonstrate that each module of AGPC is effective and that the network proposed in this paper is able to achieve good detection accuracy on multiple datasets.

\section{Related Work}
\label{section:related work}

\begin{figure}[htbp]
    \centering
    \includegraphics[width=\textwidth/2]{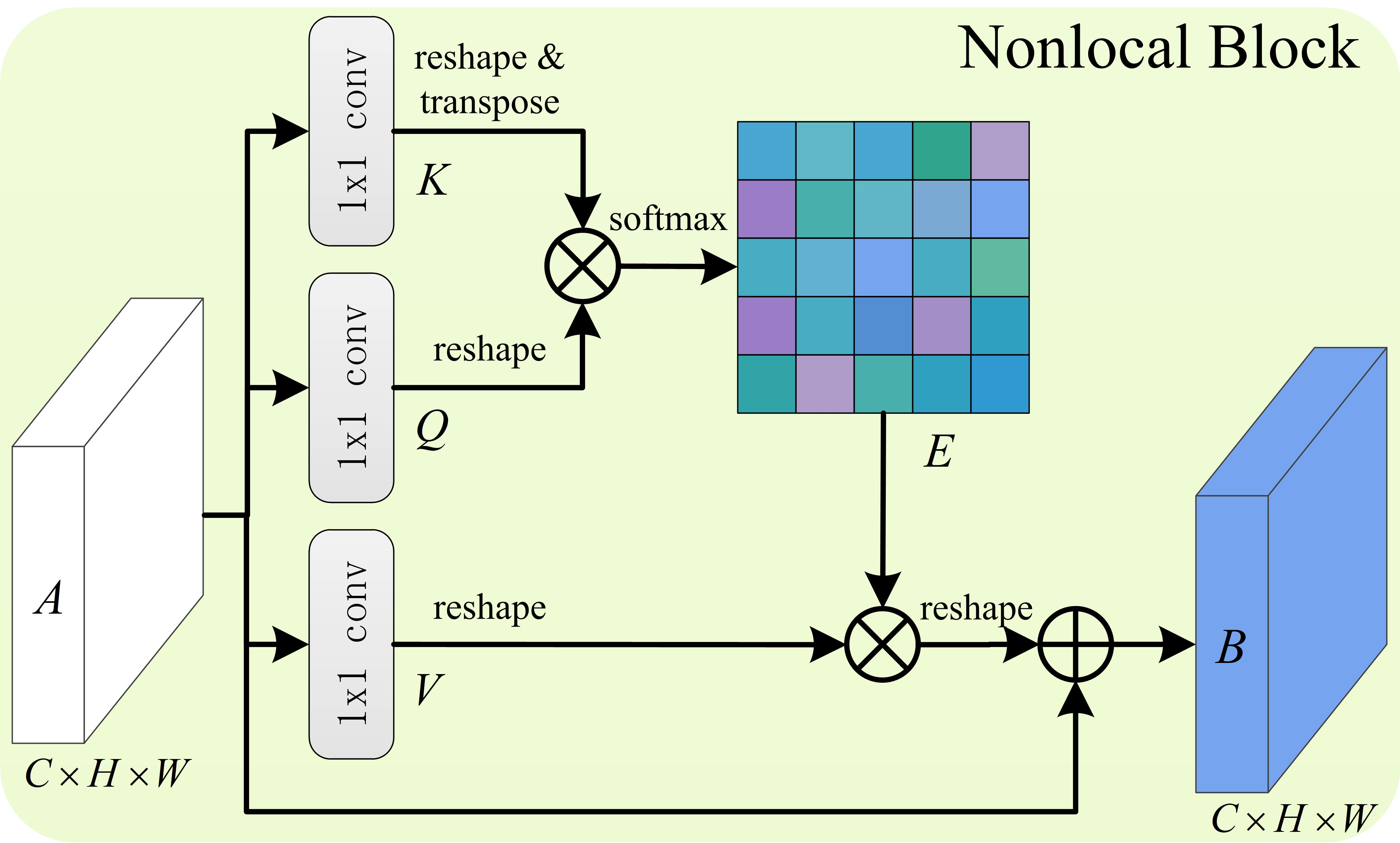}
    \caption{Illustration of nonlocal network.}
    \label{fig:nonlocal}
\end{figure}

\textbf{Context Modules.} Context modules have received a lot of attention with the presentation of Nonlocal Network\cite{wang2018non}. Inspired by nonlocal mean filtering, it uses nonlocal block shown in Fig.\ref{fig:nonlocal} to calculate the correlation between each pixel and other pixels. Because the global computation is performed the limitations of convolutional layers' field of perception are broken and long-range dependencies between pixels can be captured. As a key component, nonlocal block takes as input a feature map of size $A \in {\mathbb{R}^{C \times H \times W}}$, which is reduced by $1 \times 1$ convolution to improve computational efficiency. Then the similarity of embedding space is estimated by the embedded Gaussian function as shown in Eq.\ref{eq:nonlocal_1}, where $N=H \times W$ represents the number of positions. Finally the mapping results are added to the original feature map $A$ as shown in Eq.\ref{eq:nonlocal_2}, using scalar $\alpha$ to learn the ratio of the two adaptively to obtain the Nonlocal result $B \in {\mathbb{R}^{C \times H \times W}}$. This module can be easily embedded into neural networks and effectively enhances the performance.

\begin{equation}
    e_{ij} = \frac{exp(K_i \cdot Q_j)}{\sum_{i=1}^{N}exp(K_i \cdot Q_j)}
    \label{eq:nonlocal_1}
\end{equation}

\begin{equation}
    B_j = \alpha \sum_{i=1}^{N}{(e_{ji}V_{i}) + A_j}
    \label{eq:nonlocal_2}
\end{equation}

Although Nonblock is able to aggregate pixels of the same category together, it still suffers from computational inefficiencies and limited aggregation capabilities. As a result, many methods have been proposed to address these drawbacks. On this basis, double attention network (DANet)\cite{fu2019dual} calculated the dependencies between channels in a similar way, and selectively aggregated the two kinds of attention. GCNet\cite{cao2019gcnet} combined nonlocal block with SCNet\cite{hu2018squeeze} to explore more effective feature expression on channel domain. Point-wise spatial attention network (PSANet)\cite{zhao2018psanet} divided long-range dependence into collect attention and distribute attention. Adaptive pyramid context network (APCNet)\cite{he2019adaptive} performed multi-scale contextual representation of deep features. CCNet\cite{huang2019ccnet} calculated context dependence of the cross, and then realized global association through recurrent operation. Object context network (OCNet)\cite{yuan2018ocnet} divided deep features into blocks according to different scales and calculated the self-attention in the blocks, and achieved more robust effect.

\textbf{Attention Mechanism.} Attention mechanism is widely used in image segmentation, target detection and other fields due to its excellent performance. SEnet\cite{hu2018squeeze} selects the optimal feature map by weighting the sum of channels, CBAM\cite{woo2018cbam} combines channel attention and spatial attention to obtain good enhancement, and RACN\cite{zhang2018image} combines residual network and channel attention to obtain good reconstruction for image super resolution. SKnet\cite{li2019selective} enhances the representation of channel features by selecting kernels with multiple branches.

Context modules and attention mechanism in neural networks develops rapidly, but they are all aimed at segmentation of extended targets, most of which occupy a large area in images. However, the detection of small targets in infrared scenes is very different from previous applications. The lack of colour information, high noise, small targets area and lack of texture information all highlight the difficulty of this research. Therefore, how to develop contextual modules and attention mechanism in infrared small targets detection is very interest problem.

\textbf{Infrared Small Target Detection Networks.} In fact, research on small target detection networks has been conducted for a long time\cite{kampffmeyer2016semantic, bosquet2018stdnet, gong2021effective}, but the vast majority of it has been for visible targets, while detecting bounding boxes in a way that differs significantly from infrared small targets detection. However, With the release of public datasets in recent years, the task of infrared small targets detection is seen as a target pixel-level segmentation task\cite{long2015fully, ronneberger2015u, lin2017feature} with the necessary conditions to move towards neural networks. At the same time, it is about to move away from the previous slow development\cite{wang2019detection, zhao2020novel}. At the same time some excellent networks, such as MDFA\cite{wang2019miss} and ACM\cite{dai2021asymmetric}, have come into view. Both approach the detection task from the perspective of generative adversarial networks and feature fusion respectively.

\section{Attention-Guided Pyramid Context Network}
\label{section:agpc}

In this section, we first describe the network architecture, followed by local and global associations for AGCB and GCA contained therein, respectively. Then the multi-scale contextual pyramid is described. Finally, asymmetric fusion module is shown for feature fusion.

\subsection{Network Architecture}
\label{subsection:network architecture}

\begin{figure}[htbp]
    \centering
    \includegraphics[width=88mm]{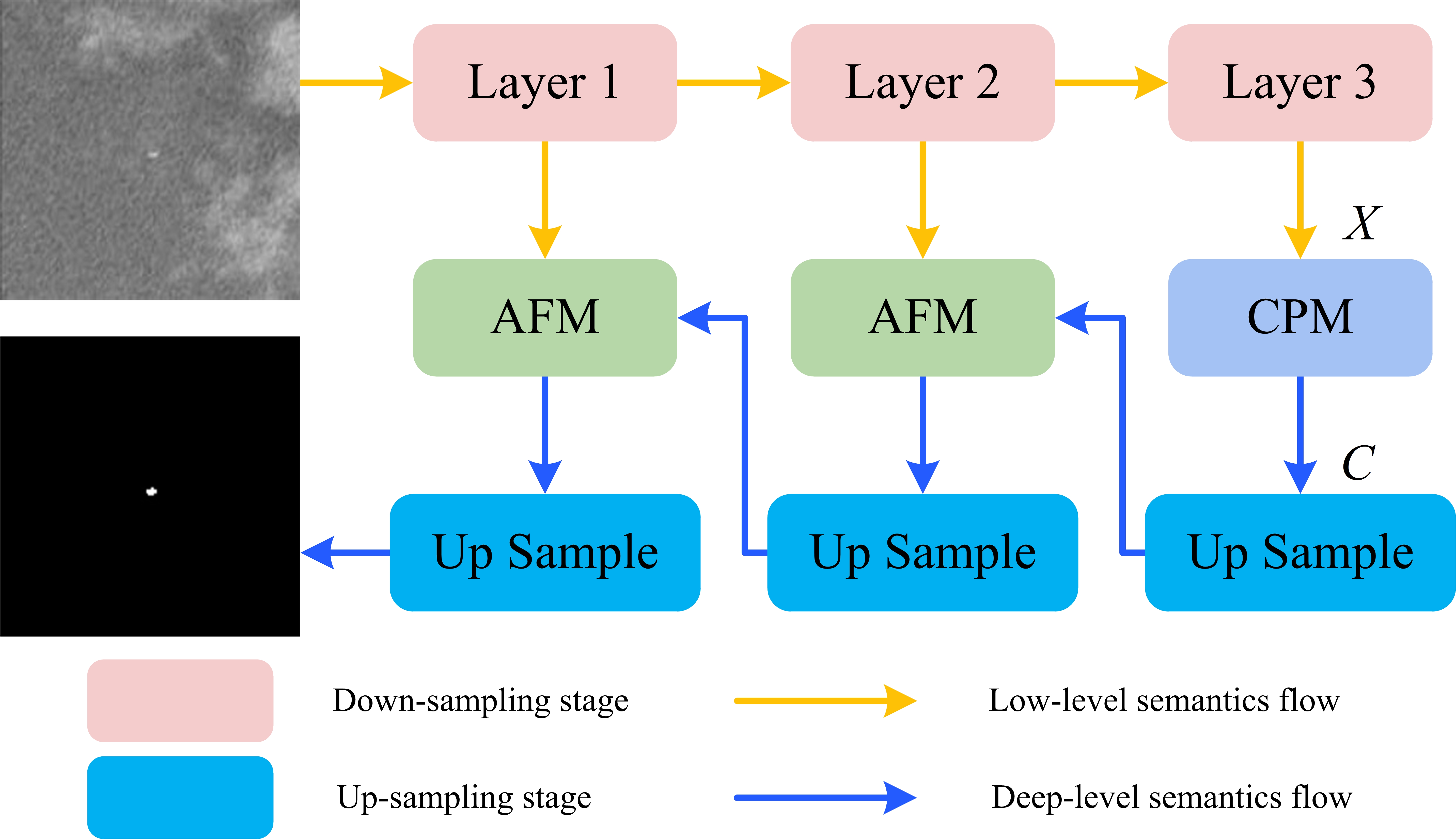}
    \caption{Overview of the proposed AGPCNet for infrared small targets detection.}
    \label{fig:overall}
\end{figure}

Given an input image $I$, we want to classify each pixel to discriminate whether it is a target or not through end-to-end processing of the neural network, and ultimately output a segmentation result of the same size as $I$. Our approach takes the infrared image $I$ and feeds it into a fully convolutional neural network\cite{he2016deep}, where it is split into three downsampling layers to obtain a feature map $X \in {\mathbb{R}^{W \times H}}$. Then let $X$ pass through the context pyramid module to generate the feature map $C$ after information aggregation. Deep-level and low-level semantics are then fused in upsampling stages through asymmetric fusion module to obtain a more accurate target localisation. The final output binary map is the infrared small target detection result. The network architecture is given in Fig.\ref{fig:overall}. 

\subsection{Attention-Guided Context Block}
\label{subsection:agcb}

\begin{figure}[htbp] 
    \centering
    \includegraphics[width=\textwidth/2]{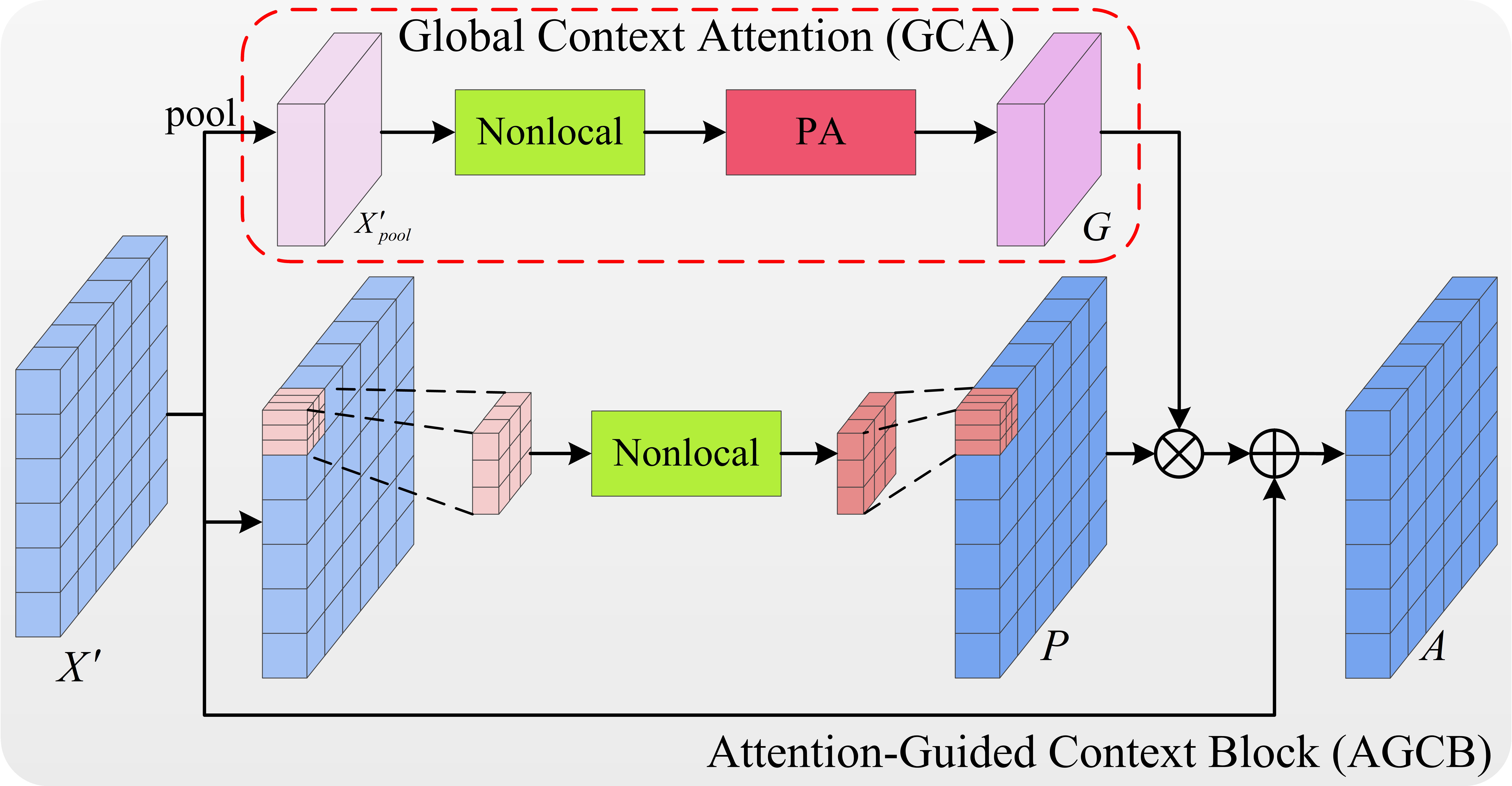}
    \caption{Illustration of Attention-Guided Context Block.}
    \label{fig:agcb}
\end{figure}

AGCB is the base module of the network as shown in Fig.\ref{fig:agcb}. Its lower branch and upper branch represent the local and global association of semantics respectively. 

Local association means dividing the feature map $X'$ into $s \times s$ patches of size $w \times h$, where $w = ceil(\frac{W}{s})$, $h=ceil(\frac{H}{s})$, and calculating the dependencies of pixels in the local range by nonlocal operation, where all patches share weights. Subsequently, the output feature maps are concentrated together to form a new local associated feature map $P \in {\mathbb{R}^{W \times H}}$. The main purpose of this is to limit the perceptual field of the network to a local range, and to use the dependencies between pixels in a local range to aggregate pixels belonging to the same category to calculate the likelihood of target occurrence. This enables the discriminative results to be obtained for local areas and to exclude the influence of structural noise within the patch on the targets. At the same time, the computation of local association can also save computational resources and speed up network training and inference.

On the other hand, at the global range, Gaussian noise in the background patch and targets in the target patch may both have similar responses in terms of local association. Therefore, we also expect the network to be able to use the global association information to assist in discriminating the target locations, to know where the real target is by aggregating features between patches, and to exclude similar patch and noise interference. Therefore, we propose an attention-guided global context module, as shown in upper branch in Fig.\ref{fig:agcb}.

Given an input feature map $X' \in {\mathbb{R}^{W \times H}}$, GCA first extracts the features of each patch by adaptive pooling and get pooled features of size $s \times s$, with each pixel inside representing the features of each patch. The contextual information between each patch are then analysed via nonlocal block. Subsequently, to integrate information between channels and obtain more accurate attentional guidance, we used pixel attention to reduce the dimensionality by a factor of 4 and then up-dimension to the original value, where pixel attention will be described in Section\ref{subsection:afm}. The results are treated as attention guide to obtain the guide map $G \in {\mathbb{R}^{s \times s}}$.

\begin{equation}
    A_{p} = \beta \cdot \delta(W[ P_1 G_1, P_2 G_2, \cdots, P_{s^2} G_{s^2} ]) + X'
    \label{eq:patch-wise}
\end{equation}

\begin{equation}
    A_{e} = \beta \cdot \delta(WP \otimes I(G)) + X'
    \label{eq:element-wise}
\end{equation}

Considering the way in which the guide map $G \in {\mathbb{R}^{s \times s}}$ is imposed on local association feature $P \in {\mathbb{R}^{W \times H}}$, two solutions are given. The first is \textbf{Patch-wise GCA}, as shown in Eq.\ref{eq:patch-wise}, which directly multiplies the elements in $G$ with each block of $P$. Another is \textbf{Pixel-wise GCA}, as shown in Eq.\ref{eq:element-wise}, which upsamples $G$ to $H \times W$ using interpolation, indicated by $I(\cdot)$, and subsequently multiplies it with the corresponding element of $P$. $P_i$ and $G_i$ denote the $i$ patch of the feature map $P$ and the $i$ element of $G$ respectively, and $\delta$ denotes the activation function, here ReLU is used. There is such a comparison mainly to consider that the attention of edges between patches may change suddenly, and interpolation can smooth the edges and get natural guide map. For a more efficient representation, we set a learning parameter $\beta$ to add output to input, using the network to self-adapt to select more efficient segmentation features.

\subsection{Context Pyramid Module}
\label{subsection:cpm}

\begin{figure}[htbp] 
    \centering
    \includegraphics[width=\textwidth/2]{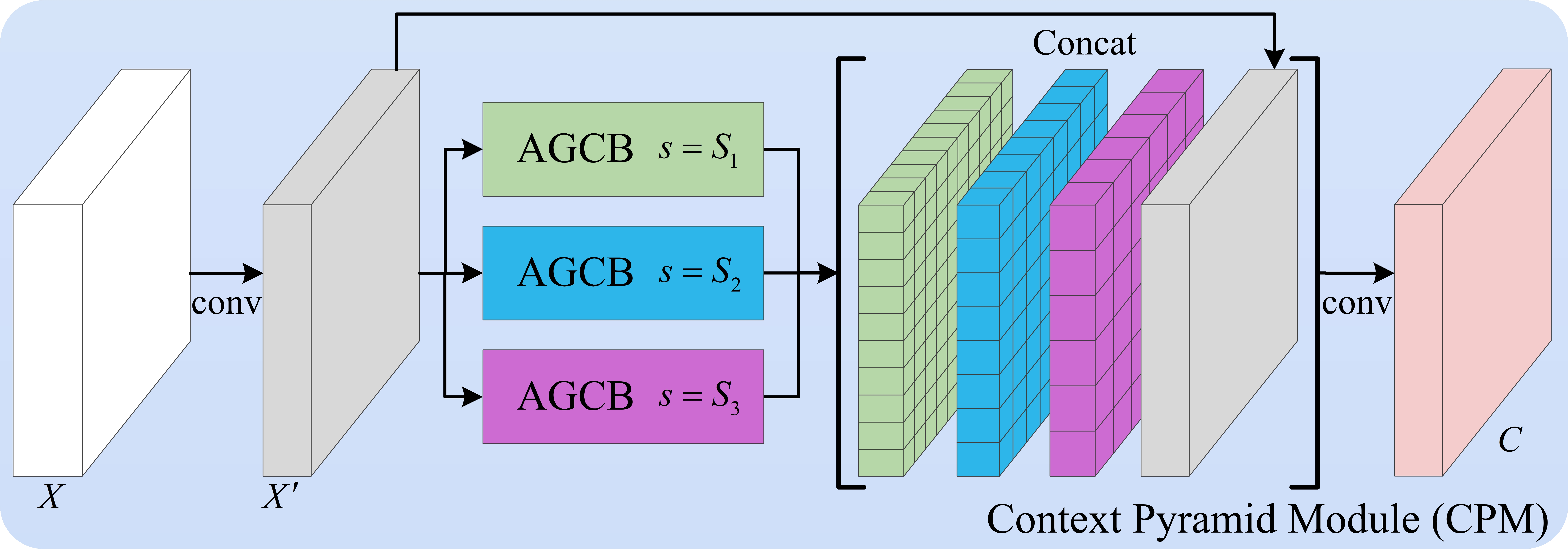}
    \caption{Illustration of Context Pyramid Module.}
    \label{fig:cpm}
\end{figure}

In the following we describe the proposed Context Pyramid Module for infrared small targets detection, the structure of which is given in Fig.\ref{fig:cpm}. The input feature maps $X$ are fed into multiple AGCBs of varying scales in parallel, after being reduced in dimensionality by $1 \times 1$ convolution, the results are denoted by $A=\{A^{S_1}, A^{S_2}, \cdots\}$, where $S$ is a vector of scales. Then the multiple aggregation feature maps $\{A^i\}$ are concentrated together with the original feature map. Finally, the channel information is integrated by $1 \times 1$ convolution, which is the output result of CPM, so that the AGCBs of different scales form a context pyramid.

\subsection{Asymmetric Fusion Module}
\label{subsection:afm}

\begin{figure}[htbp]
    \centering
    \includegraphics[width=88mm]{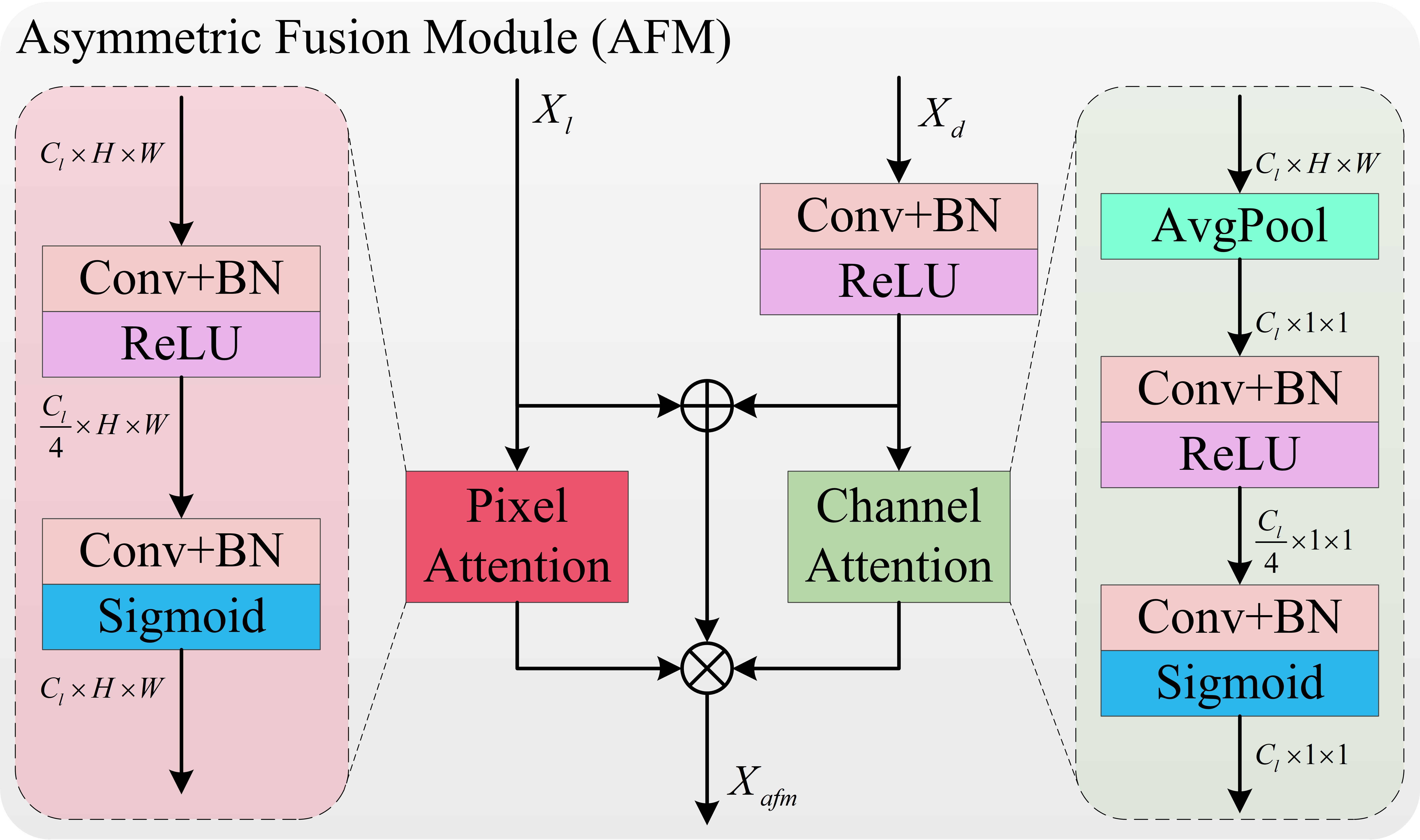}
    \caption{Illustration of Asymmetric Fusion Module.}
    \label{fig:afm}
\end{figure}

In the perspective of feature fusion, we propose a novel Asymmetric Fusion Module to merge low-level and deep-level semantics, inspired by CBAM\cite{woo2018cbam} and ACM\cite{dai2021asymmetric}. As shown in Fig.\ref{fig:afm}, the low-level semantics $X_l$ and the deep-level semantics $X_d$ are taken as input, and we do separate processing for the different information categories they contain. 

\begin{equation}
    g_{pa}(X) = \sigma(W_{2}\delta(W_{1}X))
    \label{eq:pixel-attention}
\end{equation}

\begin{equation}
    g_{ca}(X) = \sigma(W_{2}\delta(W_{1}P(X)))
    \label{eq:channel-attention}
\end{equation}

The low-level semantics $X_l$ contains a large amount of target location information and we use the point-attention mechanism\cite{dai2021asymmetric} as in Eq.\ref{eq:pixel-attention}. On the other hand, the deep semantic $X_d$, which first uses a $1 \times 1$ convolutional dimensionality reduction, contains much more information and we use the attention mechanism as in Eq.\ref{eq:channel-attention} to select the most important channels. $g_{pa}$ and $g_{ca}$ are constrained separately after the features are directly fused by sum, as shown in Eq.\ref{eq:afm}, which can solve the nuisance of feature mismatch caused by separate constraints. Where $\otimes$ and $\odot$ denotes the corresponding element multiplication and corresponding vector-tensor multiplication respectively, and $\sigma$ is Sigmoid function.

\begin{equation}
    X_{afm} = (X_l + \delta(WX_d)) \otimes g_{pa}(X_l) \odot g_{ca}(X_d)
    \label{eq:afm}
\end{equation}

\section{Experiments}
\label{section:experiments}

In this section, we use experiments to verify the effectiveness of AGPCNet. First, we describe the experimental setting, which includes the datasets, evaluation metrics, network implementation details and comparison methods. Then conduct ablation studies for each module of the network to verify the practicability of them. Finally, the visual and numerical comparison between AGPCNet and state-of-the-art methods further demonstrates that AGPCNet can accurately detection infrared small targets.

\subsection{Experimental Setting}
\label{subsection:experimental setting}
\subsubsection{Datasets And Evaluation Metrics}
\label{subsubsection:dataset and metrics}

In this article, we use both datasets published in MDvsFA cGan\cite{wang2019miss} and SIRST\cite{dai2021asymmetric}. However, there are the following problems: 1) The image sizes of the two datasets are different and do not match infrared data in actual scene. It is difficult to unify the two for one network; 2) SIRST has only 427 images in total. A small amount of data can lead to problems such as unstable network training, easy over-fitting, and failure of the model to converge. 

\begin{table*}[t]
    \renewcommand\arraystretch{1.4}
    \begin{center}
    \caption{Parameter settings of the comparison methods.}
    \label{table:parameter setting}
    \resizebox{120mm}{!}{
        \begin{tabular}{c c}
        \hline
        Methods  & Parameter Setting \\
        \hline
        Tophat \cite{tom1993morphology}      & Structure shape: disk, Size: $5 \times 5$\\
        
        LCM \cite{chen2013local}             & Filter radius: 1,2,3,4 \\
        
        MPCM \cite{wei2016multiscale}        & Filter radius: 1,2,3,4 \\
        
        NRAM \cite{zhang2018infrarednram}    & Patch size: $30 \times 30$, Slide step: 10, $\lambda = 1 / \sqrt{\min \left( m, n \right)}$ \\
        
        PSTNN \cite{zhang2019infraredpstnn}  & Patch size: $40 \times 40$, Slide step: 40, $\lambda = 0.6 / \sqrt{\max \left( n_1, n_2 \right) * n_3 }$ \\
        
        SRWS \cite{zhang2021infraredsrws}    & Patch size: $50 \times 50$, Slide step: 10, $\beta = 1 / \sqrt{\min \left( m, n \right)}$ \\
        \hline
        \end{tabular}}
    \end{center}
\end{table*}

In order to deal with these problems, we first fixed the input size of datasets and network to $256 \times 256$, which is closer to the size of actual infrared data, and can minimize the redundant design of networks. In addition, we perform data augmentation on SIRST to cope with the problem of its lack of data. For the training set, we resize each image to $512 \times 512$ according to original ratio, and then crop target patch to a size of $256 \times 256$. Crop five images for each target, and ensure that target is at the top-left, bottom-left, top-right, bottom-right, and center of the cropped image. Then each cropped image is randomly rotated at $0^\circ$, $45^\circ$, $90^\circ$, $135^\circ$, and $180^\circ$, and the angle fluctuates randomly within a certain range to simulate a real and varied scene. For the test set, in order to ensure that it is closer to real scenes, we only perform random cropping at five positions. In this way, we get training set of 8525 images and test set of 545 images, which are enough to train neural networks. In order to facilitate the research of scholars, we publish the SIRST Aug dataset to \url{https://github.com/Tianfang-Zhang/SIRST-Aug}.

In terms of evaluation metrics, we use classic semantic segmentation evaluation metrics such as $Precision$, $Recall$, $Fmeasure$, and mean Intersection over Union ($mIoU$). Their equations are shown in Eq.\ref{eq:fmeasure}, \ref{eq:miou}. $Precision$ and $Recall$ respectively represent the ratio of correctly classified pixels to all labeled target and predicted targets, and the two influence each other. In order to measure the relationship between them, we use $Fmeasure$, which means that they are equally important. The network must detect targets and ensure that there are as few false alarms as possible.



\begin{equation}
    Fmeasure = \frac{2*Precsion*Recall}{Precision+Recall}
    \label{eq:fmeasure}
\end{equation}

\begin{equation}
    mIoU = \frac{\text{\# Area of Overlap}}{\text{\# Area of Union}}
    \label{eq:miou}
\end{equation}

In addition, we also use receiver operating characteristic (ROC) curve to characterize the dynamic relationship between false postive rate ($FPR$) and true postive rate ($TPR$), and their equations are shown in Eq.\ref{eq:fpr tpr}. At the same time, area under the curve (AUC) has also been used as a key metric for quantitative evaluation of ROC.

\begin{equation}
    FPR = \frac{\sum FP}{\sum FP+TN}, \quad TPR = \frac{\sum TP}{\sum TP+FN}
    \label{eq:fpr tpr}
\end{equation}

\subsubsection{Implementation Details And Comparison Methods}
\label{subsubsection:implementation details and methods}

We implement our method based on Pytorch. The optimizer uses stochastic gradient descent (SGD), where momentum and weight decay coefficients are set to 0.9 and 0.0004 respectively. The initial learning rate is 0.05, and the attenuation strategy of poly is used at the same time, that means the learning rate is multiplied by $\left( 1 - \frac{iter}{total\_iter} \right) ^ {0.9}$ after each iteration. Batch size is set to 8, the loss function of each model is SoftIoU \cite{rahman2016optimizing} and is trained for 5 epochs. On the hardware, we use one 1080ti GPU for training.

This paper selects 8 classic comparison methods from different types to compare with AGPCNet. In model-driven schemes, BS-based method uses Tophat\cite{tom1993morphology}, HVS-based methods uses the pioneering LCM\cite{chen2013local} and MPCM\cite{wei2016multiscale} with significant effects, and optimization-based methods uses matrix optimized NRAM\cite{zhang2018infrarednram}, multi-subspace optimized SRWS\cite{zhang2021infraredsrws}, and tensor optimized PSTNN\cite{zhang2019infraredpstnn}. In data-driven schemes, we use MDvsFA cGAN\cite{wang2019miss} and ACM\cite{dai2021asymmetric} as comparison methods. The parameter settings involved in these methods are given in Table\ref{table:parameter setting}. In addition, because the result of model-driven schemes are saliency maps, binary segmentation is needed for further evaluation. We adopt a unified segmentation principle for each result and the threshold is $T = 0.5 \times max\_value$.

\subsection{Ablation Study}
\label{subsection:ablation study}

In order to verify the rationality of AGPCNet, we conduct ablation experiments with different settings.

\textbf{The effect of Backbone, CPM and AFM.} As shown in Table\ref{table:ablation whole network}, we verify their effectiveness by adding CPM and AFM to different backbones. For the backbone, to prevent overfitting due to the large amount of parameters, we used ResNet-18 and ResNet-34, and used pretrained weights. Among them, the maximum value of each backbone is marked with \textbf{bold black}, and the maximum value of each column is marked with \textcolor{red}{\textbf{bold red}}. From Table\ref{table:ablation whole network}, we can see that under the same kind of backbone, adding AFM and CPM respectively, the performance can be improved in both datasets. On different backbone, the maximum value tends to occur on the deeper ResNet-34. This indicates that deepening the network backbone does give a partial performance boost to the network.

\begin{table}[htbp]
    \renewcommand\arraystretch{1.3}
    \begin{center}
    \caption{Ablation study on the whole network.}

    \label{table:ablation whole network}
    \resizebox{90mm}{!}{
        \begin{tabular}{c|cc|cc|cc}
        
        \Xhline{1.3pt}
        \multirow{2}{*}{Backbone} & \multirow{2}{*}{CPM} & \multirow{2}{*}{AFM} & \multicolumn{2}{c|}{MDFA} & \multicolumn{2}{c}{SIRST Aug} \\
        \cline{4-7}
        & & & mIoU & Fmeasure & mIoU & Fmeasure \\
        \Xhline{1.3pt}

        ResNet-18 &         &         & 0.4304 & 0.6018 & 0.6776 & 0.8078 \\
        ResNet-18 & $\surd$ &         & 0.4487 & 0.6195 & 0.6957 & 0.8120 \\
        ResNet-18 &         & $\surd$ & 0.4556 & 0.6260 & 0.6887 & 0.8157 \\
        ResNet-18 & $\surd$ & $\surd$ & \textcolor{red}{\textbf{0.4641}} & \textcolor{red}{\textbf{0.6340}} & \textbf{0.7037} & \textbf{0.8261} \\
        \hline
        ResNet-34 &         &         & 0.4450 & 0.6159 & 0.6878 & 0.8151 \\
        ResNet-34 & $\surd$ &         & 0.4548 & 0.6252 & 0.6973 & 0.8217 \\
        ResNet-34 &         & $\surd$ & 0.4533 & 0.6238 & 0.6905 & 0.8169 \\
        ResNet-34 & $\surd$ & $\surd$ & \textbf{0.4585} & \textbf{0.6287} & \textcolor{red}{\textbf{0.7117}} & \textcolor{red}{\textbf{0.8315}} \\

        \Xhline{1.3pt}
        \end{tabular}}
    \end{center}
\end{table}

\textbf{The effect of reduction ratios.} In the network, dimensionality reduction not only reduces redundant information but also greatly speeds up the network training inference, however this may lead to information loss. In AGPCNet, we have two dimensionality drops in CPM and Nonlocal Block respectively, and their reduction ratios are denoted as $(r_c, r_n)$. We explore the most beneficial way of segmenting infrard small targets by using different dimensionality reduction ratios. As shown in Table\ref{table:ablation reduction ratio}, we limited the total ratio to 64 and 128, which means $r_c*r_n = 64,128$. Although the parameters fluctuated somewhat across different datasets, the overall variation was not significant. We can pick more profitable parameter settings during training.

\begin{table}[htbp]
    \renewcommand\arraystretch{1.3}
    \begin{center}
    \caption{Ablation study on reduction ratios.}
    \label{table:ablation reduction ratio}
    \resizebox{88mm}{!}{
        \begin{tabular}{c|cc|cc}
        \Xhline{1.3pt}
        \multirow{2}{*}{Reduction Ratios} & \multicolumn{2}{c|}{MDFA} & \multicolumn{2}{c}{SIRST Aug} \\
        \cline{2-5}
        & mIoU & Fmeasure & mIoU & Fmeasure \\
        \Xhline{1.3pt}
        (4,16) & 0.4571 & 0.6274 & \textcolor{red}{\textbf{0.7083}} & \textcolor{red}{\textbf{0.8292}} \\
        (8,8)  & 0.4590 & 0.6292 & 0.6758 & 0.8065 \\
        (16,4) & \textcolor{red}{\textbf{0.4713}} & \textcolor{red}{\textbf{0.6406}} & 0.7037 & 0.8261 \\
        \hline
        (8,16) & \textbf{0.4706} & \textbf{0.6400} & 0.6993 & 0.8231 \\
        (16,8) & 0.4468 & 0.6176 & \textbf{0.7050} & \textbf{0.8270} \\
        \Xhline{1.3pt}
        \end{tabular}}
    \end{center}
\end{table}

\textbf{The effect of GCA type and PA.} In AGPCNet we use GCA to measure the dependencies between patches as a way to achieve global associations between pixels. We discuss here the types of GCA mentioned above, namely Patch-Wise and Pixel-Wise. on the other hand, to verify the correctness of using pixel attention in GCA, we also experiment from this aspect. From Table\ref{table:ablation GCA} we can observe that the network performs the worst without GCA, and that in each substructure the absence of pixel attention, denote by PA, leads to poorer network performance, also illustrating the predominance of PA for attentional guidance. In terms of GCA type, the two are similar, with Patch-Wise being relatively higher.

\begin{table}[htbp]
    \renewcommand\arraystretch{1.3}
    \begin{center}
    \caption{Ablation study on GCA type.}
    \label{table:ablation GCA}
    \resizebox{88mm}{!}{
        \begin{tabular}{c|c|cc|cc}
        \Xhline{1.3pt}
        \multirow{2}{*}{GCA Type} & \multirow{2}{*}{PA} & \multicolumn{2}{c|}{MDFA} & \multicolumn{2}{c}{SIRST Aug} \\
        \cline{3-6}
        & & mIoU & Fmeasure & mIoU & Fmeasure \\
        \Xhline{1.3pt}

        None         &         & 0.4312 & 0.6026 & 0.6908 & 0.8171 \\
        Patch-Wise   &         & 0.4458 & 0.6167 & 0.7021 & 0.8250 \\ 
        Patch-Wise   & $\surd$ & \textcolor{red}{\textbf{0.4672}} & \textcolor{red}{\textbf{0.6368}} & \textcolor{red}{\textbf{0.7300}} & \textcolor{red}{\textbf{0.8443}} \\ 
        \hline
        Pixel-Wise   &         & 0.4557 & 0.6261 & 0.6960 & 0.8208 \\
        Pixel-Wise   & $\surd$ & \textbf{0.4631} & \textbf{0.6330} & \textbf{0.7250} & \textbf{0.8406} \\
        
        \Xhline{1.3pt}
        \end{tabular}}
    \end{center}
\end{table}

\textbf{The effect of patch size in CPM.} We analyse the scale or rather the size of each patch in the CPM as a key parameter. Since the image in the backbone is downsampled 8 times, the feature map size is obtained as $32 \times 32$, i.e. $W=H=32$. This means that the pixels of each feature map represent the features of an 8*8 region on the original image. As shown in Table\ref{table:ablation scale}, we set up multiple sets of comparison experiments, and it should be noted that the parameter here is the patch size $w=h$, which corresponds to the scale $s=ceil(\frac{W}{w})$. For the selection of patch sizes, we wanted to cover both the target area and a certain background area, therefore experiments were carried out on both single sizes and multi-scale combinations. As can be seen from the table, the single size experiments incorporated relatively little information, resulting in poorer results. In contrast, the larger the number of patch sizes combined in CPM, the better results tend to be. The best results for both datasets occur for combinations of four sizes.

\begin{table}[htbp]
    \renewcommand\arraystretch{1.3}
    \begin{center}
    \caption{Ablation study on scales.}
    \label{table:ablation scale}
    \resizebox{80mm}{!}{
        \begin{tabular}{c|cc|cc}
            \Xhline{1.3pt}
        \multirow{2}{*}{Patch Size} & \multicolumn{2}{c|}{MDFA} & \multicolumn{2}{c}{SIRST Aug} \\
        \cline{2-5}
        & mIoU & Fmeasure & mIoU & Fmeasure \\
        \Xhline{1.3pt}

        \{3\}        & 0.4480 & 0.6188 & 0.6663 & 0.7997 \\
        \{5\}        & 0.4454 & 0.6163 & 0.6794 & 0.8091 \\
        \{6\}        & 0.4513 & 0.6219 & 0.7100 & 0.8304 \\
        \hline
        \{3,5\}      & 0.4569 & 0.6272 & 0.6919 & 0.8179 \\
        \{3,6\}      & 0.4535 & 0.6240 & 0.6960 & 0.8208 \\
        \hline
        \{3,5,6\}    & 0.4601 & 0.6302 & 0.6947 & 0.8199 \\
        \{3,5,8\}    & 0.4573 & 0.6276 & \textbf{0.7192} & \textbf{0.8367} \\
        \hline
        \{3,5,6,8\}  & \textbf{0.4694} & \textbf{0.6389} & \textcolor{red}{\textbf{0.7272}} & \textcolor{red}{\textbf{0.8420}} \\
        \{3,5,6,10\} & \textcolor{red}{\textbf{0.4727}} & \textcolor{red}{\textbf{0.6419}} & 0.6884 & 0.8155 \\
        \Xhline{1.3pt}
        \end{tabular}}
    \end{center}
\end{table}

\begin{figure*}[htbp]
    \centering
    \includegraphics[width=140mm]{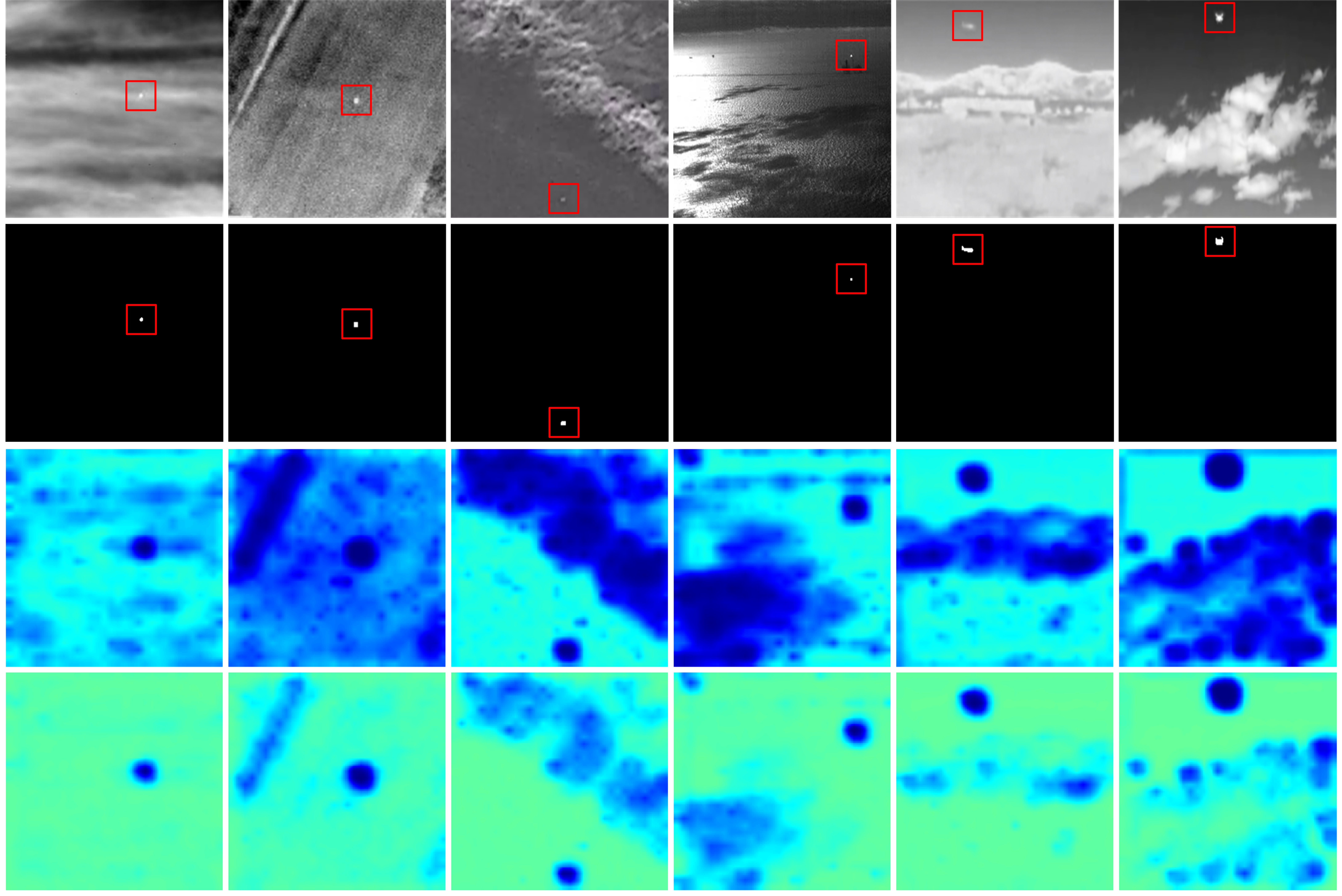}
    \caption{Illustration of heatmap. The rows from top to bottom are Infrared Images, Groundtruth, Feature Map $X$ input into CPM, Heatmap $C$ output from CPM.}
    \label{fig:heatmap}
\end{figure*}

In addition, to demonstrate that CPM really does focus network attention on the target area, we use six scenarios for this purpose. As shown in Fig.\ref{fig:heatmap}, the rows from top to bottom represent the original image, groundtruth, feature map $X$ input into the CPM, and heatmap map $C$ output from the CPM. The target region is marked with a red box and the color from light to dark illustrates the energy distribution from sparse to dense. We can observe that the energy of feature map $X$ is distributed throughout the complex background of the image, which inevitably causes interference in the accurate detection of small targets. On the heatmap $C$ output from CPM, on the other hand, the energy is much more concentrated in the target area and the background clutter is much less influential. This also illustrates CPM's ability to perceive contextual information in both local and global associations, focusing the attention on the target region.

\textbf{The effect of PA and CA in AFM.} In addition, as a key module, we also analyzed the impact of pixel attention and channel attention in AFM. As shown in Table\ref{table:ablation AFM}, we used ResNet-18 as the backbone and added two separate modules for comparison. It can be seen that both have a boosting effect on the network, and that working with both together brings optimal results. It also shows that AFM can efficiently fuse low-level and deep-level semantics.

\begin{table}[htbp]
    \renewcommand\arraystretch{1.3}
    \begin{center}
    \caption{Ablation study on AFM.}
    \resizebox{88mm}{!}{
    \label{table:ablation AFM}
        \begin{tabular}{c|cc|cc|cc}
        \Xhline{1.3pt}
        \multirow{2}{*}{Backbone} & \multirow{2}{*}{PA} & \multirow{2}{*}{CA} & \multicolumn{2}{c|}{MDFA} & \multicolumn{2}{c}{SIRST Aug} \\
        \cline{4-7}
        & & & mIoU & Fmeasure & mIoU & Fmeasure \\
        \Xhline{1.3pt}

        ResNet-18 &         &         & 0.4329 & 0.6042 & 0.6812 & 0.8104 \\
        ResNet-18 & $\surd$ &         & 0.4439 & 0.6149 & 0.7011 & 0.8243 \\
        ResNet-18 &         & $\surd$ & 0.4492 & 0.6199 & 0.7005 & 0.8239 \\
        ResNet-18 & $\surd$ & $\surd$ & \textcolor{red}{\textbf{0.4668}} & \textcolor{red}{\textbf{0.6365}} & \textcolor{red}{\textbf{0.7071}} & \textcolor{red}{\textbf{0.8284}} \\
        
        \Xhline{1.3pt}
        \end{tabular}}
    \end{center}
\end{table}

\subsection{Comparison to State-of-the-Art Methods}
\label{subsection:comparsion}

In the previous section, we proved the effectiveness of each module in AGPCNet, and also conducted experimental analysis on the values of parameters. In this section, we compare AGPCNet with state-of-the-art methods through visual and numerical evaluation to further verify the effectiveness of our method.

\subsubsection{Visual Evaluation}
\label{subsubsection:visual eval}

\begin{figure*}[htbp]
    \centering
    \includegraphics[width=140mm]{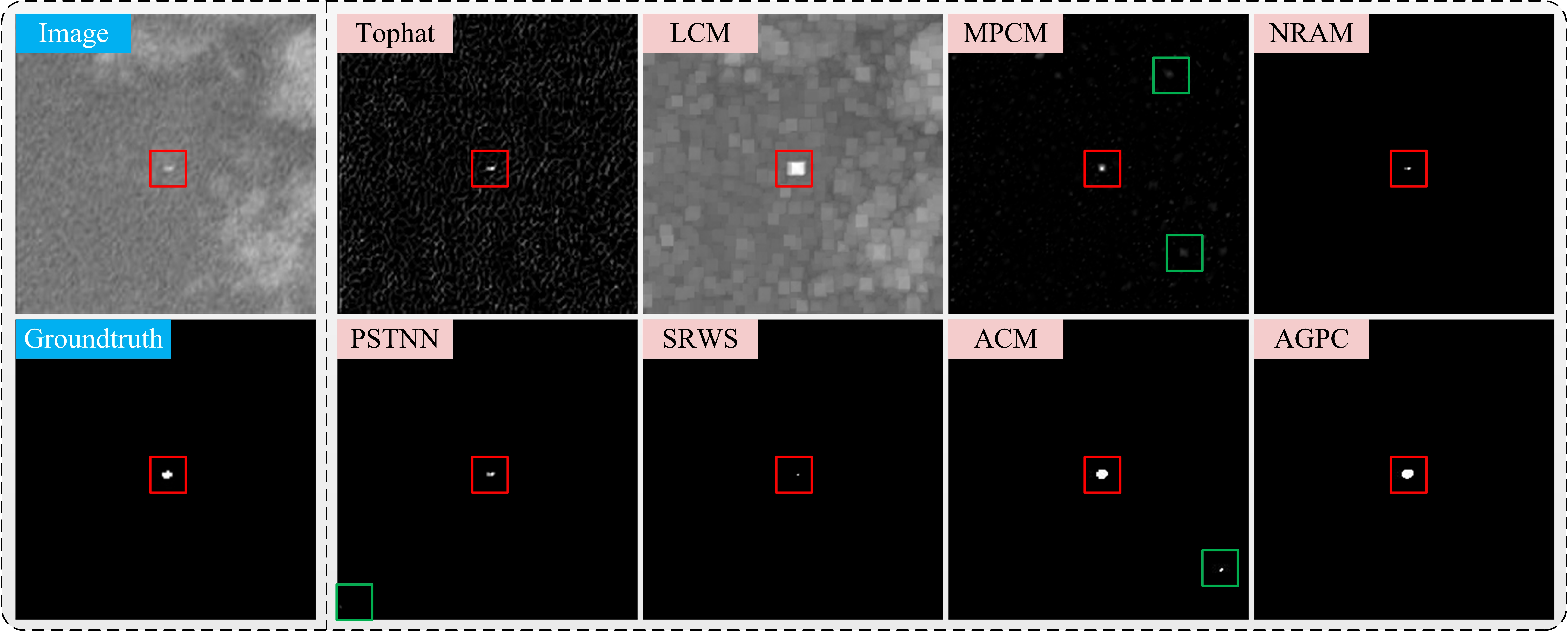}
    \caption{Illustration of infrared scene 1.}
    \label{fig:result1}
\end{figure*}

\begin{figure*}[htbp]
    \centering
    \includegraphics[width=140mm]{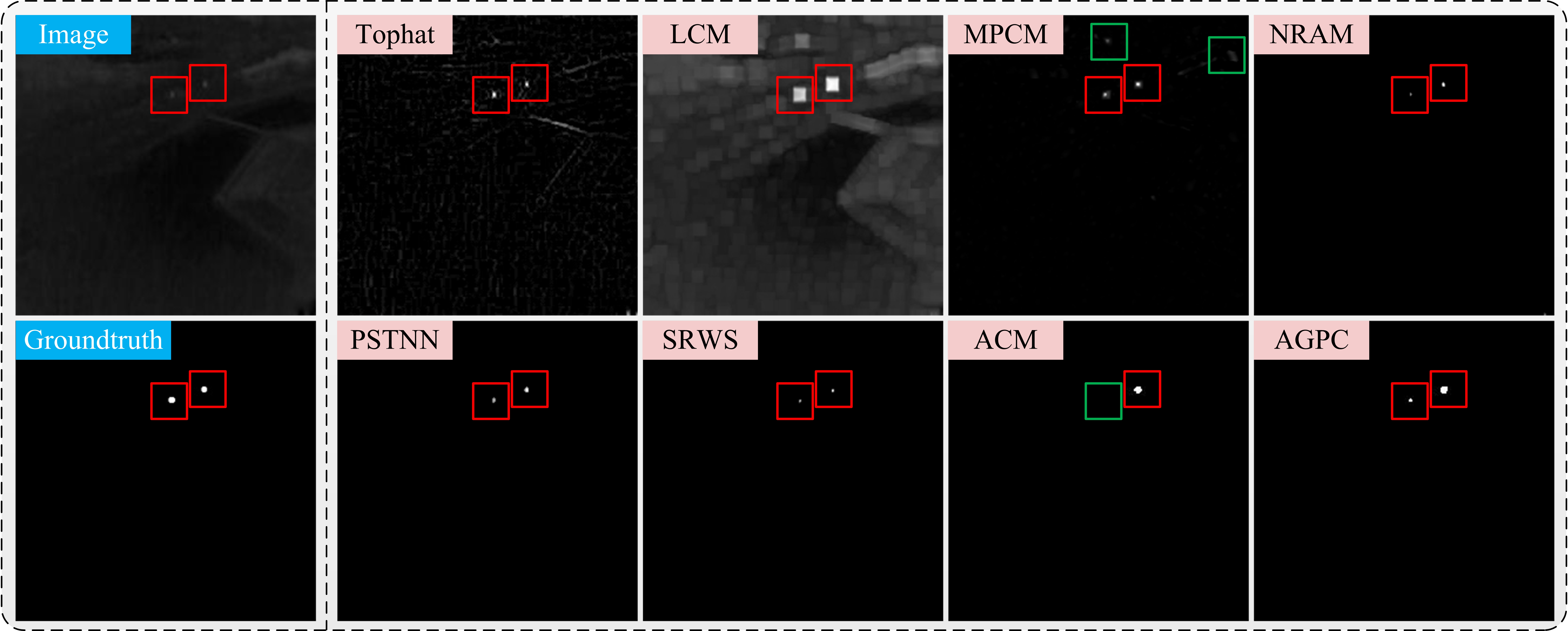}
    \caption{Illustration of infrared scene 2.}
    \label{fig:result2}
\end{figure*}

\begin{figure*}[htbp]
    \centering
    \includegraphics[width=140mm]{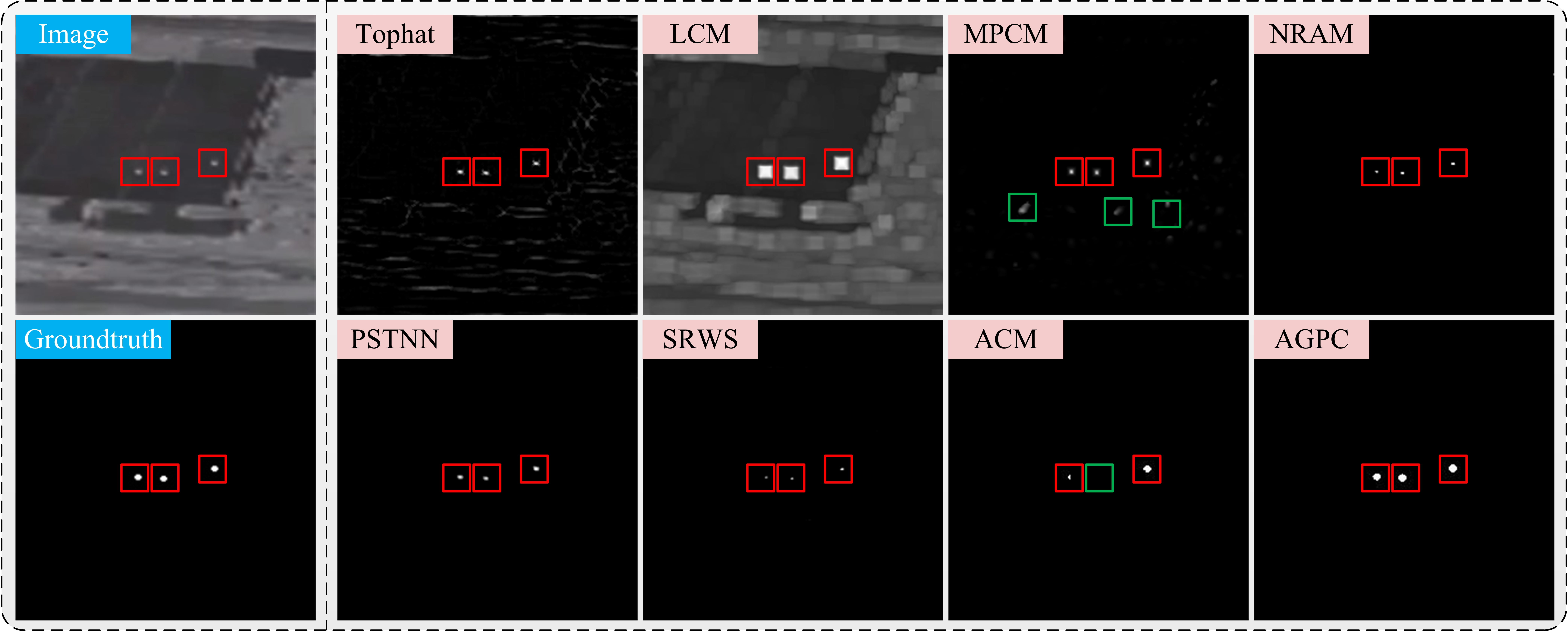}
    \caption{Illustration of infrared scene 3.}
    \label{fig:result3}
\end{figure*}

\begin{figure*}[htbp]
    \centering
    \includegraphics[width=140mm]{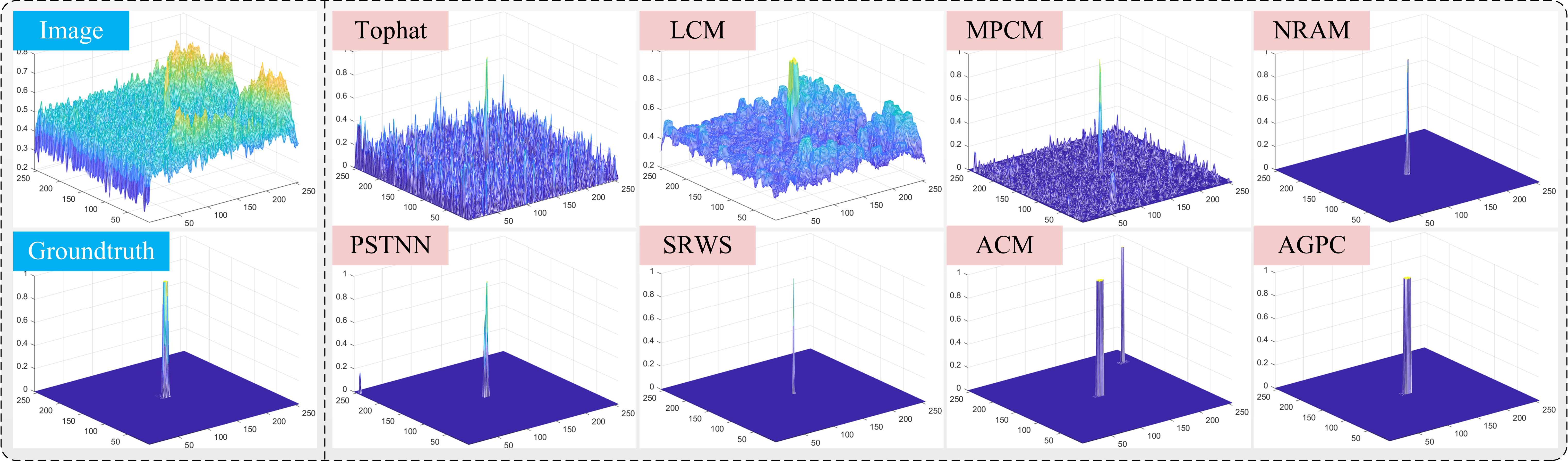}
    \caption{3D display of infrared scene 1.}
    \label{fig:result1_3d}
\end{figure*}

\begin{figure*}[htbp]
    \centering
    \includegraphics[width=140mm]{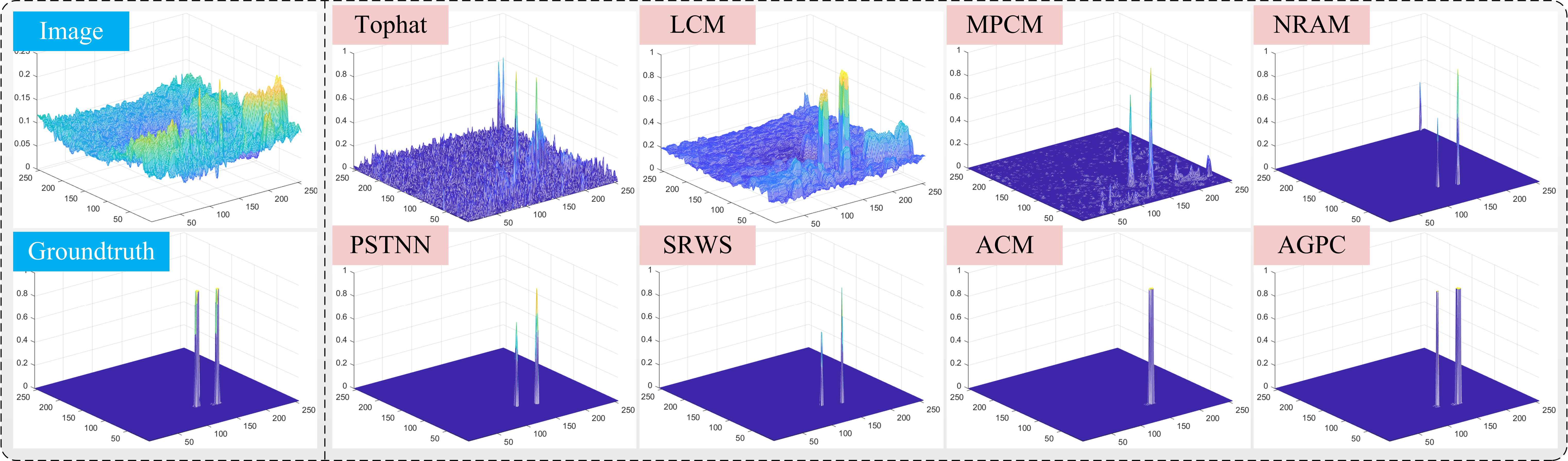}
    \caption{3D display of infrared scene 2.}
    \label{fig:result2_3d}
\end{figure*}

\begin{figure*}[htbp]
    \centering
    \includegraphics[width=140mm]{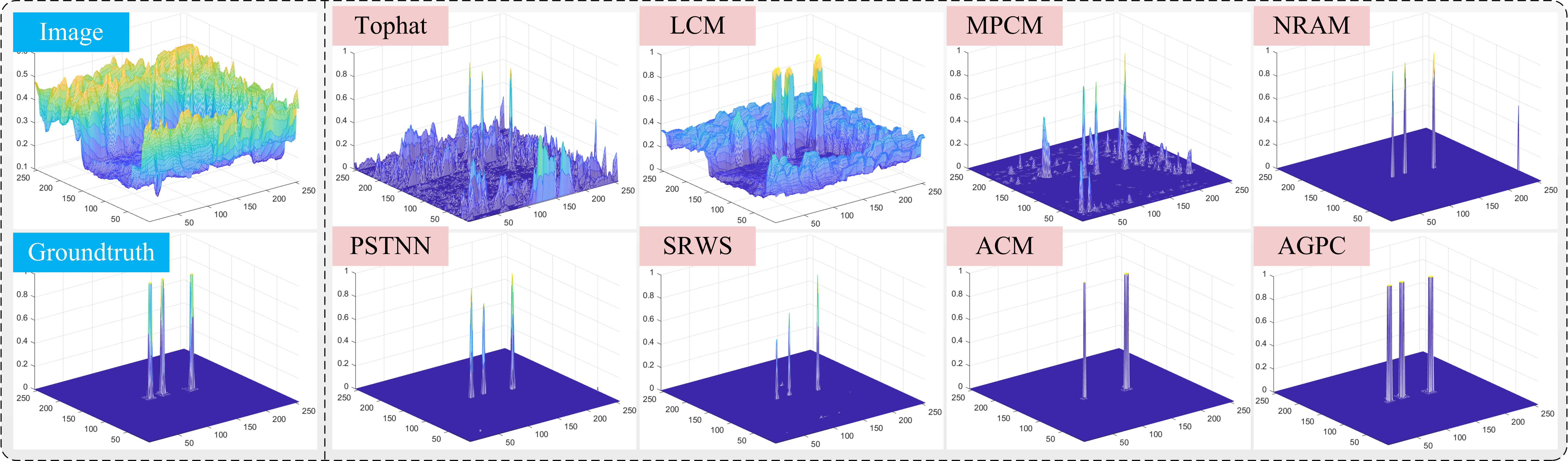}
    \caption{3D display of infrared scene 3.}
    \label{fig:result3_3d}
\end{figure*}

As shown in Fig.\ref{fig:result1}-\ref{fig:result3}, we selected three typical infrared small target scenes and compared the detection results of 8 methods. In the figures, we use red boxes to mark the target position, and green boxes to mark imperceptible false alarms and miss detections. In order to facilitate the observation of clutters in detection results, Fig.\ref{fig:result1_3d}-\ref{fig:result3_3d} is the 3D display of three scenes. The detection method is annotated in the upper left corner of each image.

From the figures, we can observe that BS-based Tophat method is sensitive to noise, and its detection results have strong response to noise and edge clutter in background. In HVS-based methods, the detection results of LCM is not satisfactory. And although MPCM can detect targets, a large amount of clutter remains in background. This is because the background assumption of HVS is relatively simple, and it is difficult to distinguish between the two using global information when background and targets are similar. The three optimization-based methods can accurately detect targets. But compared with groundtruth, it can be seen that they can only perceive the approximate location. Due to the addition of various constraints guided by prior knowledge, they have strict constraints on targets. This also leads to a small target area segmented in the results. It can also be seen from the figures that these methods that rely heavily on models, assumptions and parameters are not robust.

In data-driven approaches, because ACM only integrates low-level semantics and deep-level semantics, and does not consider the association between targets and global information, miss detection occurs in scene 1, and false alarms occur in scene 2, 3. This also illustrates the importance of contextual information. Due to the support of CPM and AFM, AGPCNet can not only accurately detect targets, but also segment targets region as completely as possible.

\subsubsection{Numerical Evaluation}
\label{subsubsection:numerical eval}

In order to accurately illustrate the effectiveness of AGPCNet, we use numerical methods for quantitative evaluation in this section. Table\ref{table:comparison SOTA} quantifies the various metrics of the comparison methods on the MDFA and SIRST Aug datasets. The maximum value of each column is marked in \textcolor{red}{\textbf{bold red}}, and the second largest value is marked in \textbf{bold black}. It is worth mentioning that because the image input size of MDvsFA cGan is $128 \times 128$, which is smaller than the input size of this paper, it is difficult to reproduce, so the experimental results of this method shown here are intercepted from paper \cite{wang2019miss}. At the same time, the input size of ACM is $480 \times 480$, which is also different from $256 \times 256$ in this paper. We removed the maxpooling layer in the first down-sampling stage, and used the size and datasets in this paper to retrain ACM.

From Table\ref{table:comparison SOTA}, we can see that, except for LCM, all model-driven methods exhibit high $Precision$ and low $Recall$. This is because previous classic infrared small target detection methods pay more attention to detecting targets position, while ignoring the importance of complete segmentation of targets. These methods often do not detect the entire region of targets, but few pixels in the target region, which can also be found in Fig.\ref{fig:result1}-\ref{fig:result3}. This leads to a very high $Precision$ of these methods. On the contrary, $Recall$ represents the ratio of accurately segmented pixels to all labeled pixels. A low $Recall$ indicates that these methods have only detected a few pixels in real targets.

The reason for this phenomenon is that in model-driven approaches, the inevitable key problem is to suppress background. Then various constraints based on prior knowledge are introduced, including noise L21 norm\cite{zhang2018infrarednram}, structure tensor\cite{zhang2019infraredpstnn}, self-regularization item\cite{zhang2021infraredsrws}. And these constraints will inevitably produce response at targets region. In other words, the model-driven approaches suppressed targets while suppressing background, which leads to targets detected by this category of approaches occupies few pixels.

In data-driven approaches, $Recall$ is relatively high. Since they are not constrained by prior knowledge, the result of networks depends on targets and background characteristics it has learned. They are making a balance between accurate segmentation and reduction of false alarms. We can see that AGPCNet is the best on both $mIoU$ and $Fmeasure$ and has a relatively high improvement.

\begin{table*}[!htbp]
    \renewcommand\arraystretch{1.4}
    \begin{center}
    \caption{Comparison with state-of-the-art methods on MDFA and SIRST Aug.}

    \label{table:comparison SOTA}
        \begin{tabular}{c|cc:ccc|cc:ccc}
            \Xhline{1.3pt}
        \multirow{2}{*}{Methods} & \multicolumn{5}{c|}{MDFA} & \multicolumn{5}{c}{SIRST Aug} \\
        \cline{2-11}
        & \multicolumn{1}{c}{Precision} & \multicolumn{1}{c}{Recall} & \multicolumn{1}{c}{mIoU} & \multicolumn{1}{c}{Fmeasure} & \multicolumn{1}{c|}{AUC} & \multicolumn{1}{c}{Precision} & \multicolumn{1}{c}{Recall} & \multicolumn{1}{c}{mIoU} & \multicolumn{1}{c}{Fmeasure} & \multicolumn{1}{c}{AUC} \\
        \Xhline{1.3pt}

        Tophat      & 0.0399 & 0.1046 & 0.0298 & 0.0578 & 0.7154 & 0.6873 & 0.0818 & 0.0788 & 0.1461 & 0.7441 \\
        LCM         & 0.0114 & 0.7525 & 0.0113 & 0.0224 & \textcolor{red}{\textbf{0.9512}} & 0.0182 & 0.7943 & 0.0182 & 0.0357 & 0.8016 \\
        MPCM        & 0.6472 & 0.1499 & 0.1386 & 0.2434 & 0.8476 & 0.9357 & 0.1509 & 0.1493 & 0.2599 & 0.7886 \\
        NRAM        & 0.7398 & 0.1064 & 0.1026 & 0.1861 & 0.4981 & 0.8830 & 0.0813 & 0.0804 & 0.1489 & 0.5104 \\
        PSTNN       & 0.7236 & 0.1544 & 0.1458 & 0.2545 & 0.6087 & 0.9366 & 0.0949 & 0.0943 & 0.1724 & 0.3556 \\
        SRWS        & 0.8138 & 0.0394 & 0.0390 & 0.0751 & 0.4641 & 0.8418 & 0.0360 & 0.0357 & 0.0690 & 0.5706 \\
        MDvsFA cGAN & 0.6600 & 0.5400 & -      & 0.6000 & \textbf{0.9100} & -      & -      & -      & -    & -  \\
        ACM         & 0.5454 & 0.7006 & \textbf{0.4423} & \textbf{0.6133} & 0.8688 & 0.8223 & 0.8317 & \textbf{0.7051} & \textbf{0.8270} & \textbf{0.9283} \\
        AGPCNet        & 0.5939 & 0.7241 & \textcolor{red}{\textbf{0.4843}} & \textcolor{red}{\textbf{0.6525}} & 0.8682 & 0.8323 & 0.8542 & \textcolor{red}{\textbf{0.7288}} & \textcolor{red}{\textbf{0.8431}} & \textcolor{red}{\textbf{0.9344}} \\
        
        \Xhline{1.3pt}
        \end{tabular}
    \end{center}
\end{table*}

\begin{figure*}[!h]
    \centering
    \includegraphics[width=150mm]{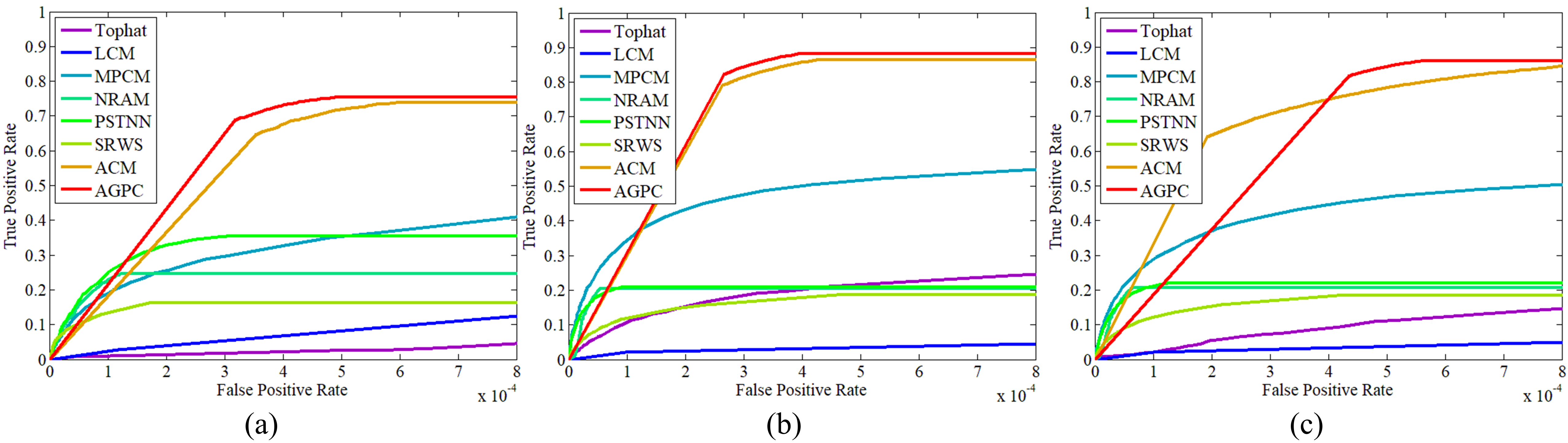}
    \caption{Illustration of ROC compared with state-of-the-art methods.}
    \label{fig:roc}
\end{figure*}

As mentioned in Section\ref{subsubsection:dataset and metrics}, we have unified the input size of the MDFA and SIRST Aug datasets, which is also convenient for doing some interesting work. We merged the train set and test set of the two datasets together, and get a merged dataset with 18525 images in training set and 645 images in test set. This can also eliminate the differences caused by the labeling style as much as possible. 

As shown in Table\ref{table:comparison SOTA merged}, we conduct experiments on this merged dataset. It can be seen that AGPCNet has reached the maximum value on almost all the metrics. In addition, in order to show the comparison of AUC more vividly, Fig.\ref{fig:roc} shows ROC curves of these methods. These experimental data shows that AGPCNet can greatly suppress background, accurately detect targets, and segment targets more accurately than other state-of-the-art methods.

\begin{table}[htbp]
    \renewcommand\arraystretch{1.3}
    \begin{center}
    \caption{Comparison with state-of-the-art methods on the merged dataset.}
    \label{table:comparison SOTA merged}
        \begin{tabular}{c|cc:ccc}
        \Xhline{1.3pt}
        \multirow{2}{*}{Methods} & \multicolumn{5}{c}{Merged} \\
        \cline{2-6}
        & \multicolumn{1}{c}{Precision} & \multicolumn{1}{c}{Recall} & \multicolumn{1}{c}{mIoU} & \multicolumn{1}{c}{Fmeasure} & \multicolumn{1}{c}{AUC} \\
        \Xhline{1.3pt}

        Tophat      & 0.2808 & 0.0834 & 0.0687 & 0.1286 & 0.7337 \\
        LCM         & 0.0175 & 0.7913 & 0.0174 & 0.0343 & 0.8256 \\
        MPCM        & 0.9071 & 0.1508 & 0.1485 & 0.2586 & 0.7897 \\
        NRAM        & 0.8677 & 0.0831 & 0.0820 & 0.1516 & 0.5098 \\
        PSTNN       & 0.9070 & 0.0992 & 0.0982 & 0.1788 & 0.3806 \\
        SRWS        & 0.8396 & 0.0362 & 0.0360 & 0.0694 & 0.5595 \\
        ACM         & 0.7203 & 0.7877 & \textbf{0.6032} & \textbf{0.7525} & \textcolor{red}{\textbf{0.9265}} \\
        Ours        & 0.7453 & 0.8384 & \textcolor{red}{\textbf{0.6517}} & \textcolor{red}{\textbf{0.7891}} & \textbf{0.9194} \\
        
        \Xhline{1.3pt}
        \end{tabular}
    \end{center}
\end{table}

In summary, model-driven approaches usually have higher $Precision$, but $Recall$ is very low. This is because there are many false alarms in the results of such methods. At the same time, they put more emphasis on detecting targets instead of accurately segmenting targets. In data-driven approaches, MDvsFA cGan used two generators for two subtasks of false alarms and false detections. Although the effect is acceptable, the network design is cumbersome and difficult to combine with other data. On another algorithm, ACM fused low-level and deep-level semantics and lacks attention to contextual information. This makes the network only perceive limited receptive field, ignoring the difference between global background context and targets.

In the experiments of this paper, we augmented the SIRST dataset for the problem of a small amount of data and unified the size of two datasets. Then the specific implementation details of AGPCNet and comparison methods are introduced. Subsequently, we verified the important role of each module separately by ablation studies, also including the selection of patch size, the role of backbone, the dimensionality reduction ratios, GCA type, etc. Finally, we compared AGPCNet with state-of-the-art methods through visual and numerical evaluation. From various experimental data, we can see that AGPCNet is an effective method that can accurately detect infrared small targets.

However, there are still some issues worthy of further investigation, such as dealing with network overfitting, more efficient contextual information, etc. In future work, we will continue to explore the application of attention mechanism in infrared small targets detection.

\section{Conclusion}
\label{section:conclusion}

In this paper, a novel infrared small targets detection method called Attention-Guided Pyramid Context Network (AGPCNet) is proposed. This learning-based and data-driven approach proposed Attention-Guided Context Block (AGCB), which divides the feature map into patches to compute local associations, and uses Global Context Attention (GCA) to compute global associations between semantics. Subsequently, from a multi-scale perspective, Context Pyramid Module (CPM) is proposed to integrate features from multi-scale AGCBs. Finally, Asymmetric Fusion Module (AFM) is proposed to integrate low-level and deep-level semantics from a feature fusion perspective to further enhance the utilization of features. In terms of dataset, the SIRST is also augmented and published to adapt to the training of neural networks. Finally, ablation studies proved the effectiveness of each module, and extensive experiments also illustrate that AGPCNet has the ability to cope with precise detection tasks of complex scenes from the perspective of visual and numerical evaluation.


 

\bibliographystyle{IEEEtran}
\bibliography{ref}

\newpage

\section{Biography Section}

\begin{IEEEbiography}[{\includegraphics[width=1in,height=1.25in,clip,keepaspectratio]{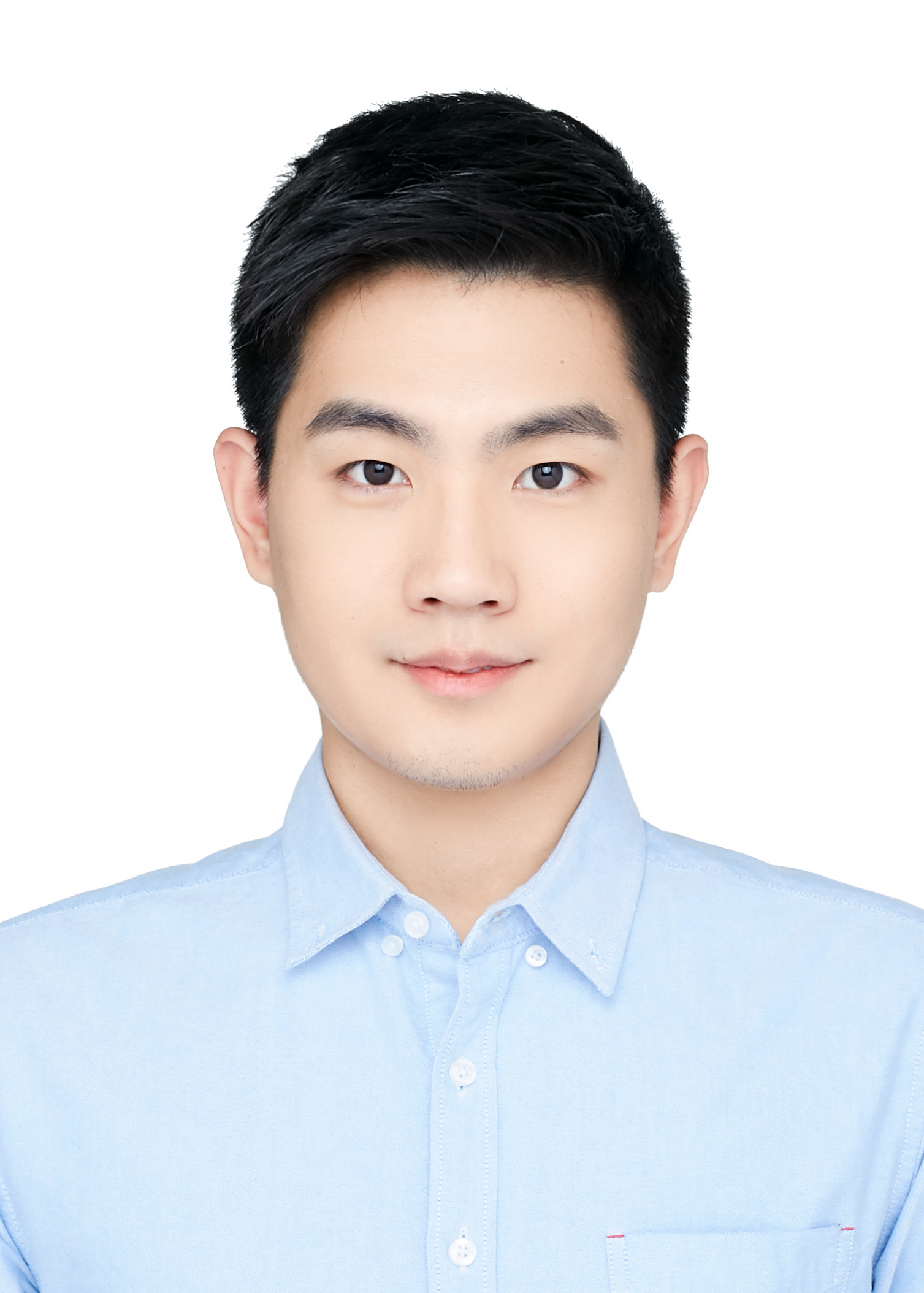}}]{Tianfang Zhang}
received his bachelor's degree from the School of Optoelectronic Information, University of Electronic Science and Technology of China (UESTC) in 2017. Currently, he is pursuing the PhD degree in School of Information and Communication Engineering, UESTC. His main research interests are infrared small target detection, compressed sensing and optimization.
\end{IEEEbiography}

\begin{IEEEbiography}[{\includegraphics[width=1in,height=1.25in,clip,keepaspectratio]{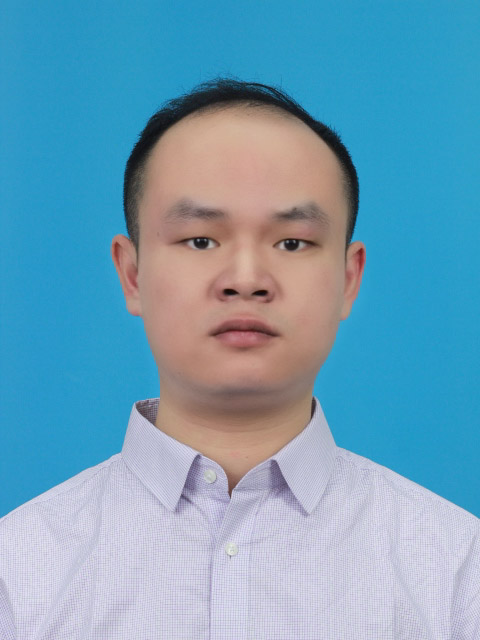}}]{Siying Cao}
received his B.S. degree from School of Optoelectronic Information, University of Electronic Science and Technology of China (UESTC) in 2013. Currently, he is pursuing the Ph.D. degree in School of Information and Communication Engineering, UESTC. His research interests include image processing, signal and information processing, and infrared target detection.
\end{IEEEbiography}

\begin{IEEEbiography}[{\includegraphics[width=1in,height=1.25in,clip,keepaspectratio]{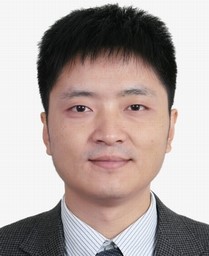}}]{Tian Pu}
received his Ph.D degree from Beijing Institute of Technology, Beijing, China, in 2002. He is currently the lecturer in School of Information and Communication Engineering, University of Electronic Science and Technology of China (UESTC). His research interests include image processing, computer vision and medical image analysis.
\end{IEEEbiography}

\begin{IEEEbiography}[{\includegraphics[width=1in,height=1.25in,clip,keepaspectratio]{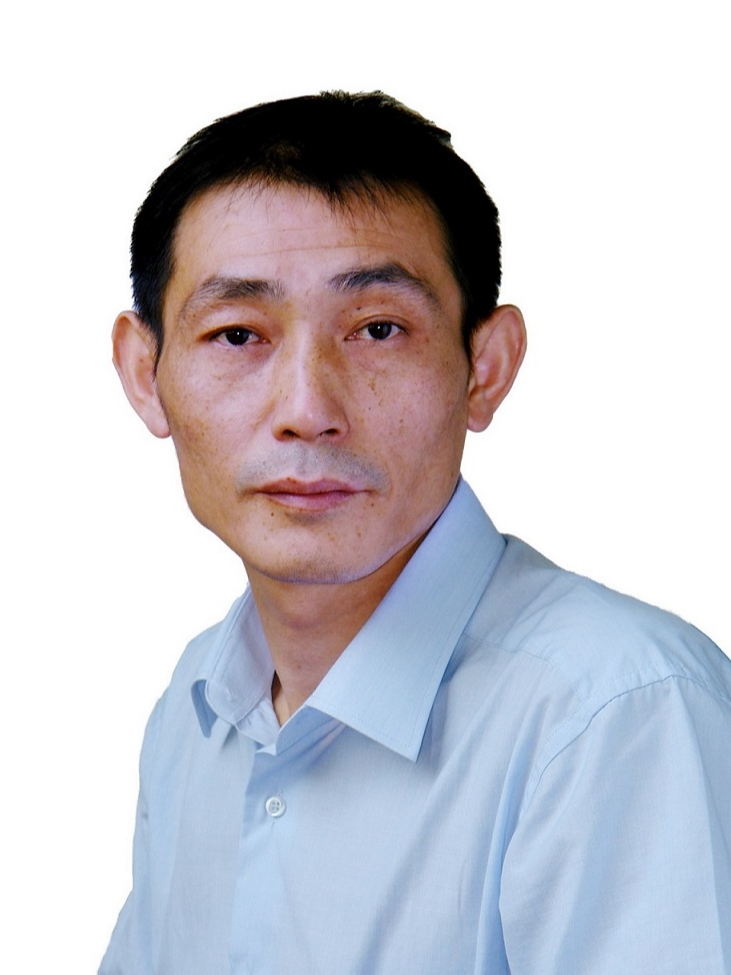}}]{Zhenming Peng}
(M’06) received his Ph.D. degree in geodetection and information technology from the Chengdu University of Technology, Chengdu, China, in 2001.
From 2001 to 2003, he was a Post-Doctoral Researcher with the Institute of Optics and Electronics, Chinese Academy of Sciences, Beijing, China. He is currently a Professor with the University of Electronic Science and Technology of China, Chengdu. His research interests include image processing, signal processing, and target recognition and tracking.
Prof. Peng is a member of Institute of Electrical and Electronics Engineers (IEEE), Optical Society of America (OSA), China Optical Engineering Society (COES), and Chinese Society of Astronautics (CSA).
\end{IEEEbiography}

\vfill

\end{document}